\documentclass[10pt,twocolumn,letterpaper]{article}

\usepackage{iccv}
\usepackage{times}
\usepackage{epsfig}
\usepackage{graphicx}
\usepackage{amsmath}
\usepackage{amssymb}

\usepackage{amsmath,amsfonts,bm}

\def\eqref#1{equation~\ref{#1}}

\def\1{\bm{1}}

\DeclareMathAlphabet{\mathsfit}{\encodingdefault}{\sfdefault}{m}{sl}
\SetMathAlphabet{\mathsfit}{bold}{\encodingdefault}{\sfdefault}{bx}{n}

\usepackage{url}
\usepackage{siunitx}
\usepackage{booktabs}
\usepackage{soul}
\usepackage{capt-of}
\usepackage[font=small]{caption}
\usepackage[breaklinks=true,bookmarks=false]{hyperref}
\usepackage{xcolor}

\newcommand{\para}[1]{\noindent\textbf{#1}:}

\iccvfinalcopy 

\def\iccvPaperID{****} 

\begin{document}

\title{Environment predictive coding for embodied agents}


\author{Santhosh K. Ramakrishnan$^{1,2}$~~~Tushar Nagarajan$^{1,2}$~~~Ziad Al-Halah$^{1}$~~~Kristen Grauman$^{1,2}$ \\
$^{1}$The University of Texas at Austin~~~~~~~~$^{2}$Facebook AI Research\\
{\footnotesize \texttt{\{srama, tushar\}@cs.utexas.edu,ziadlhlh@gmail.com,grauman@fb.com}} \\
}

\maketitle
\ificcvfinal\thispagestyle{empty}\fi

\begin{abstract}
 We introduce \emph{environment predictive coding}, a self-supervised approach to learn environment-level representations for embodied agents. In contrast to prior work on self-supervised learning for images, we aim to jointly encode a series of images gathered by an agent as it moves about in 3D environments. We learn these representations via a \emph{zone prediction task}, where we intelligently mask out portions of an agent's trajectory and predict them from the unmasked portions, conditioned on the agent's camera poses. By learning such representations on a collection of videos, we demonstrate successful transfer to multiple downstream navigation-oriented tasks. Our experiments on the photorealistic 3D environments of Gibson and Matterport3D show that our method outperforms the state-of-the-art on challenging tasks with only a limited budget of experience.
\end{abstract}

\section{Introduction}

In visual navigation tasks, an intelligent embodied agent must move around a 3D environment using its stream of egocentric
observations to sense objects and obstacles, typically without the benefit of a pre-computed map. Significant recent progress
on this problem can be attributed to the availability of large-scale visually rich 3D datasets~\cite{chang2017matterport,xia2018gibson,replica19arxiv},
developments in high-quality 3D simulators~\cite{mattersim,ai2thor,habitat19iccv,xia2020interactive}, and research on deep
memory-based architectures that combine geometry and semantics for learning representations of the 3D
world~\cite{gupta2017cognitive,henriques2018mapnet,chen2019learning,fang2019scene,chaplot2020learning,chaplot2020neural}.

Deep reinforcement learning approaches to visual navigation often suffer from sample inefficiency, overfitting,
and instability in training. Recent contributions work towards overcoming these limitations for various navigation
and planning tasks. The key ingredients are learning good image-level representations~\cite{das2018embodied,gordon_iccv19,lin_nips19,sax2020learning},
and using modular architectures that combine high-level reasoning, planning, and low-level navigation~\cite{gupta2017cognitive,chaplot2020learning,gan2019look,ramakrishnan2020occant}.

Prior work uses supervised image annotations~\cite{mirowski_arxiv16,das2018embodied,sax2020learning} and self-supervision~\cite{gordon_iccv19,lin_nips19}
to learn good image representations that are transferrable and improve sample efficiency for embodied tasks. While promising,
such learned image representations only encode the scene in the nearby locality. However, embodied agents also need higher-level
semantic and geometric representations of their history of observations, grounded in 3D space, in order to reason about the
larger environment around them.

Therefore, a key question remains: \emph{how should an agent moving through a visually rich 3D environment encode its series of egocentric observations?}
Prior navigation methods build \emph{environment-level} representations of observation sequences via memory models such as
recurrent neural networks~\cite{wijmans2019decentralized}, maps~\cite{henriques2018mapnet,chen2019learning,chaplot2020learning},
episodic memory~\cite{fang2019scene}, and topological graphs~\cite{savinov2018semi,chaplot2020neural}. However, these approaches
typically use hand-coded representations such as occupancy maps~\cite{chen2019learning,chaplot2020learning,ramakrishnan2020occant,karkus2019differentiable,gan2019look}
and semantic labels~\cite{narasimhan2020seeing,chaplot2020object}, or specialize them by learning end-to-end for solving a
specific task~\cite{wijmans2019decentralized,henriques2018mapnet,parisotto2018neural,cheng2018geometry,fang2019scene}.

\begin{figure*}
    \centering
    \vspace*{-0.3in}
    \includegraphics[width=\textwidth,trim={0 10cm 4cm 0},clip]{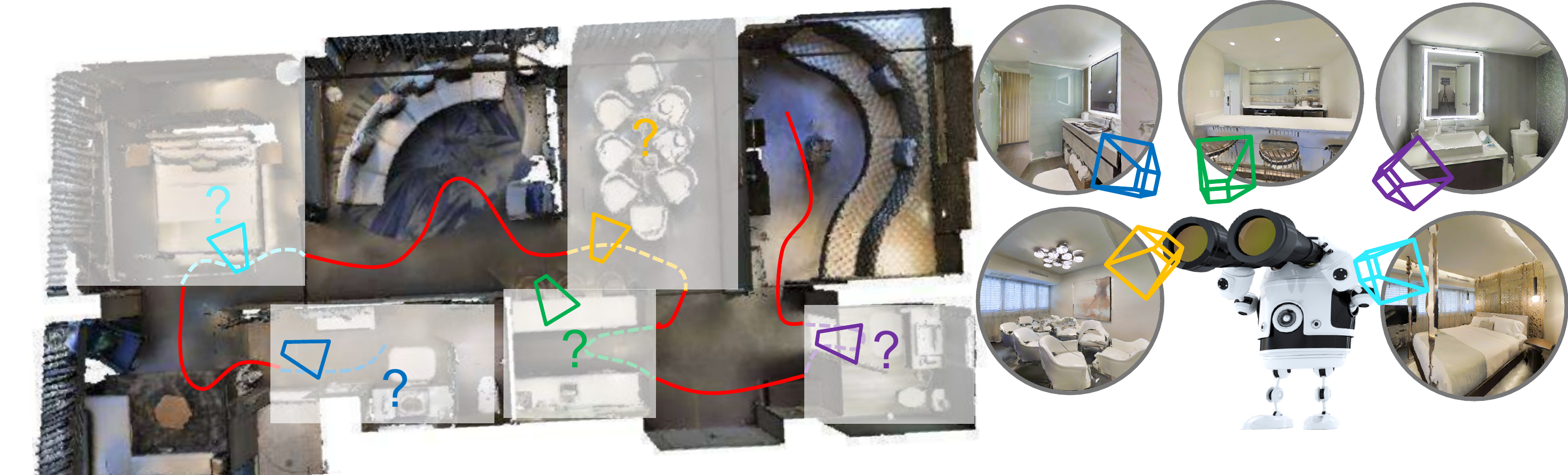}
    \caption{
        \small \textbf{Environment Predictive Coding:} During self-supervised learning, our model is given video walkthroughs of various
        3D environments. We mask portions out of the trajectory (dotted lines) and learn to infer them from the unmasked parts (in red).
        We specifically mask out all overlapping views in a local neighborhood to limit the content shared with the unmasked views.
        The resulting EPC encoder builds environment-level representations of the seen content that are predictive of the unseen content
        (marked with a ``?"), conditioned on the camera poses. The agent then uses this learned encoder in multiple navigational tasks
        in novel environments.
    }
    \label{fig:intro_figure}
    \vspace*{-0.3cm}
\end{figure*}

In this work, we introduce \emph{environment predictive coding} (EPC), a self-supervised approach to learn flexible representations
of 3D environments that are transferrable to a variety of navigation-oriented tasks. The key idea is to learn to encode a series of
egocentric observations in a 3D environment so as to be predictive of visual content that the agent has not yet observed. For example,
consider an agent that just entered the living room in an unfamiliar house and is searching for a refrigerator. It must be able to
predict where the kitchen is and reason that it is likely to contain a refrigerator. The proposed EPC model aims to learn representations
that capture these natural statistics of real-world environments in a self-supervised fashion, by watching videos recorded by other agents.
See Fig.~\ref{fig:intro_figure}.

To this end, we devise a self-supervised \emph{zone prediction} task in which the model learns  environment embeddings by watching
egocentric view sequences from other agents navigating in 3D environments in pre-collected videos. Specifically, we segment each
video into zones of visually and geometrically connected views, while ensuring limited overlap across zones in the same video.
Then, we randomly mask out zones, and predict the masked views conditioned on both the unmasked zones' views and the masked zones'
camera poses.  Intuitively, to perform this task successfully, the model needs to reason about the geometry and semantics of the
environment to figure out what is missing. We devise a transformer-based model to infer the masked visual features. Our general strategy
can be viewed as a context prediction task in sequential data~\cite{devlin2018bert,sun2019videobert,han2019video}---but, very differently,
aimed at representing high-level semantic and geometric priors in 3D environments to aid embodied agents who act in them.

Through extensive experiments on Gibson and Matterport3D, we show that our method achieves good improvements on multiple navigation-oriented
tasks compared to both state-of-the-art models and baselines that learn  image-level embeddings. 
\section{Related work}

\para{Self-supervised visual representation learning} Prior work leverages self-supervision to learn image and video
representations from large collections of unlabelled data. Image representations attempt proxy tasks such as
inpainting~\cite{pathak_cvpr16} and instance discrimination~\cite{oord_arxiv18,chen2020simple,he2020momentum},
while video representation learning leverages signals such as temporal consistency~\cite{wei2018learning,fernando2017self,kim2019self}
and contrastive predictions~\cite{han2019video,sun2019learning}. The VideoBERT project~\cite{sun2019learning,sun2019videobert}
jointly learns video and text representations from unannotated videos via filling in masked out information. Dense Predictive Coding~\cite{han2019video,han2020memory}
learns video representations that capture the slow-moving semantics in videos. Whereas these methods focus on capturing human activity for video recognition, we aim
to learn geometric and semantic cues in 3D spaces for embodied agents.  Accordingly, unlike the existing video models~\cite{sun2019learning,sun2019videobert,han2019video},
which simply infer missing frame features, our approach explicitly grounds its predictions in the 3D relationships between views.

\para{Representation learning via auxiliary tasks for RL} Reinforcement learning approaches often suffer from high sample
complexity, sparse rewards, and unstable training. Prior work tackles these challenges by using auxiliary tasks for
learning \emph{image} representations~\cite{mirowski_arxiv16,gordon_iccv19,lin_nips19,shen_iccv19,ye2020auxiliary}. In contrast,
we encode image sequences from embodied agents to obtain \emph{environment-level} representations. Recent work also learns state
representations via future prediction and implicit models~\cite{ha2018recurrent,eslami2018neural,gregor_neurips19,hafner2019dream,guo2020bootstrap}.
In particular, neural rendering approaches achieve impressive reconstructions for arbitrary viewpoints~\cite{eslami2018neural,kumar2018consistent}.
However, unlike our idea, they focus on pixelwise reconstruction, and their success has been limited to synthetically
generated environments like DeepMind Lab~\cite{dmlab}. In contrast to any of the above, we use egocentric videos to
learn predictive feature encodings of photorealistic 3D environments to capture their naturally occurring regularities.

\para{Scene completion} Past work in scene completion performs pixelwise reconstruction of 360 panoramas~\cite{dinesh2018ltla,ramakrishnan2019emergence},
image inpainting~\cite{pathak_cvpr16}, voxelwise reconstructions of 3D structures and semantics~\cite{song2016ssc}, and
image-level extrapolation of depth and semantics~\cite{song2018im2pano3d,Yang_2019_CVPR}. Recent work on visual navigation
extrapolates maps of room-types~\cite{wu2019bayesian,narasimhan2020seeing} and occupancy~\cite{ramakrishnan2020occant}.
While our approach is also motivated by anticipating unseen elements, we learn to extrapolate in a high-dimensional
feature space (rather than pixels, voxels, or semantic categories) and in a self-supervised manner without relying
on human annotations.  Further, the proposed model learns from egocentric video sequences captured by other agents,
without assuming access to detailed scans of the full 3D environment as in past work.

\para{Learning image representations for navigation} 
Prior work exploits ImageNet pretraining~\cite{gupta2017cognitive,anderson2018vision,chen2019learning}, mined object
relations~\cite{yang2019visual}, video~\cite{chang2020semantic}, and annotated datasets from various image tasks~\cite{sax2020learning,chaplot2020neural}
to aid navigation. While these methods also consider representation learning in the context of navigation tasks, they
are limited to learning image-level functions for classification and proximity prediction. In contrast, we learn predictive
representations for sequences of observations conditioned on the camera poses.
\begin{figure*}
    \centering
    \vspace*{-0.3in}
    \includegraphics[width=\textwidth,trim={0 7.5cm 1cm 0},clip]{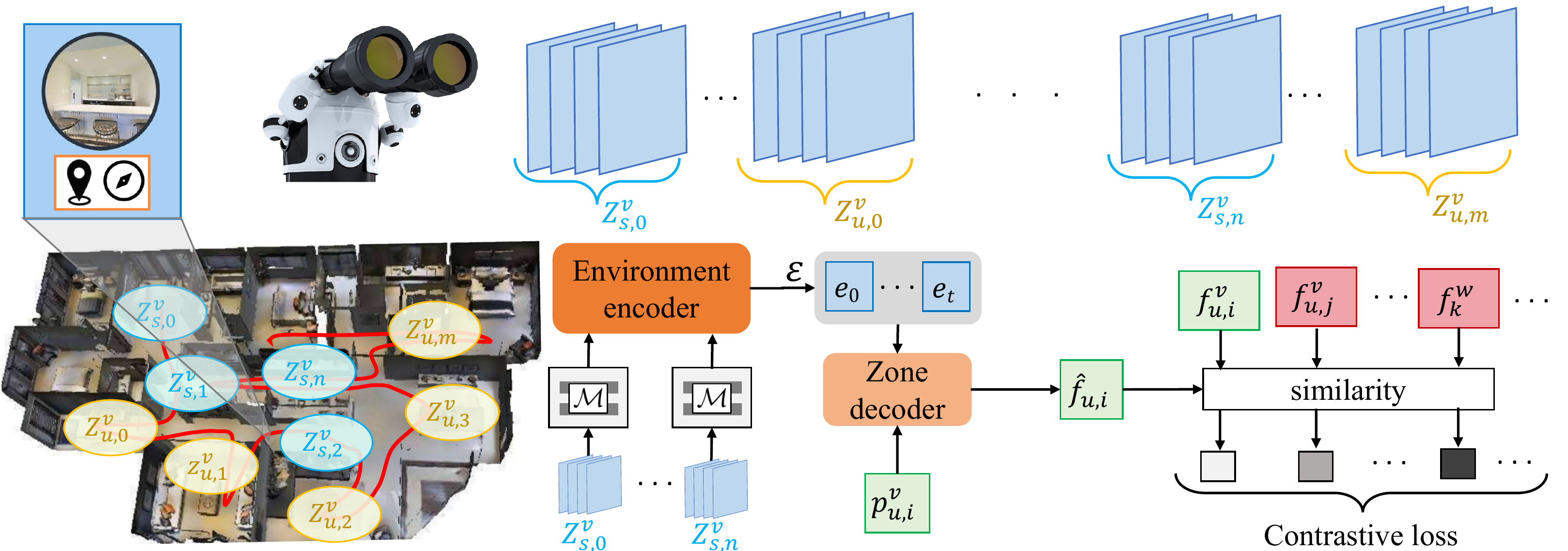}
    \caption{
        \small We propose the \emph{zone prediction task} for self-supervised learning of environment embeddings from video
        walkthroughs generated by other agents. Each frame consists of the egocentric view and camera pose (top left).
        We group the frames in video $v$ into seen zones in cyan $\{Z_{s,0}^v, \cdots, Z_{s,n}^v\}$ and unseen zones in
        yellow $\{Z_{u,0}^v, \cdots, Z_{u,m}^v\}$ (top row). The zones are generated automatically based on viewpoint overlap
        in 3D space (bottom left). Given a camera pose $p_{u,i}^v$ sampled from the unseen zone $Z_{u,i}^v$, we use a
        transformer-based encoder-decoder architecture that generates environment embeddings $\mathcal{E}$ from the seen
        zones, and predicts the feature encoding $\hat{f}_{u,i}$ of $Z_{u,i}^{v}$ conditioned on the pose $p_{u,i}^v$
        (bottom center). The model is trained to distinguish the positive $f_{u,i}^v$ from negatives in the same video
        $\{f_{u,j}^v\}_{j \neq i}$ as well from other videos $\{f_k^w\}_{t \neq s}$ (bottom right).
    }
    \label{fig:approach}
\end{figure*}

\section{Approach}

We propose \emph{environment predictive coding} (EPC) to learn self-supervised environment-level representations (Sec.~\ref{sec:epc}).
To demonstrate the utility of these representations, we integrate them into a transformer-based navigation architecture
and refine them for individual tasks (Sec.~\ref{sec:smt}). As we will show in Sec.~\ref{sec:results}, our approach leads to
both better performance and better sample efficiency compared to existing approaches.

\subsection{Environment predictive coding}\label{sec:epc}

Our hypothesis is that it is valuable for an embodied agent to learn a predictive coding of the environment.
The agent must not just encode the individual views it observes, but also learn to leverage the encoded information
to anticipate the unseen parts of the environment. Our key idea is that the environment embedding must be predictive
of unobserved content, conditioned on the agent's camera pose. This equips an agent with the structural and semantic
priors of 3D environments to quickly perform new tasks, like finding the refrigerator or covering more area.

We propose the proxy task of zone prediction to achieve this goal (see Fig.~\ref{fig:approach}). For this task,
we use a dataset of egocentric video walkthroughs collected in parallel by other agents
deployed in various unseen environments (Fig.~\ref{fig:approach}, top). For each video, we assume access
to RGB-D, egomotion data, and camera intrinsics.  Specifically, our current implementation uses egocentric camera
trajectories from photorealistic scanned indoor environments (Gibson~\cite{xia2018gibson}) to sample the training
videos; we leave leveraging in-the-wild consumer video as a challenge for future work.

We do \emph{not} assume that the agents who generated those training videos were acting to address a particular
navigation task.  In particular, their behavior need not be tied to the downstream navigation-oriented tasks for
which we test our learned representation.  For example, a training video may show agents moving about to maximize
their area coverage, or simply making naive forward-biased motions, whereas the encoder we learn is
applicable to an array of navigation tasks (as we will demonstrate in Sec.~\ref{sec:results}). Furthermore, we
 assume that the environments seen in the videos are
\emph{not} accessible for interactive training. In practice, this means that we can collect data
from different robots deployed in a large number of environments in parallel, without having to actually train our navigation
policy on those environments. These assumptions are  much weaker than those made by prior work on imitation
learning and behavioral cloning that rely on \emph{task-specific} data generated from experts~\cite{bojarski2016end,giusti2016machine}.

Our method works as follows.  First, we automatically segment videos into ``zones" which contain frames with
significant view overlaps. We then perform the self-supervised zone prediction task on the segmented videos.
Finally, we incorporate the learned environment encoder into an array of downstream navigation-oriented tasks.
We explain each step in detail next.

\vspace*{-0.1in}
\paragraph{Zone generation}

At a glance, one might first consider masking arbitrary individual frames in the training videos.  However, doing
so is inadequate for representation learning, since unmasked frames having high viewpoint overlap with the masked
frame can make its prediction trivial. Instead, our approach masks \emph{zones} of frames at once.  We define a zone
to be a set of frames in the video which share a significant overlap in their viewpoints. We also require that the
frames across multiple zones share little to no overlap.

To generate these zones, we first cluster frames in the videos based on the amount of pairwise-geometric overlap
between views. We estimate the viewpoint overlap $\psi(o_i, o_j)$ between two frames $o_{i}$, $o_{j}$ by measuring
their intersection in 3D point clouds obtained by backprojecting depth inputs into 3D space. See Appendix for more
details. For a video of length $L$, we generate a distance matrix $D \in \mathbb{R}^{L \times L}$
where $D_{i, j} = 1 - \psi(o_{i}, o_{j})$. We then perform hierarchical agglomerative clustering~\cite{lukasova1979hierarchical}
to cluster the video frames into zones based on $D$ (see Fig.~\ref{fig:approach}, bottom left). While these zones
naturally tend to overlap near their edges, they typically capture disjoint sets of content in the video. Note that
the zones segment \emph{video trajectories}, not floorplan maps, since we do not assume access to the full 3D environment.

\vspace*{-0.1in}
\paragraph{Zone prediction task}

Having segmented the video into zones, we next present our EPC zone prediction task to learn environment embeddings
(see Fig.~\ref{fig:approach}). The main motivation in this task is to infer unseen zones in the video by
previewing the \emph{global context} spanning multiple seen zones. We randomly divide the
video $v$ into seen zones $\{Z_{s,i}^{v}\}_{i=1}^{n}$ (cyan) and unseen zones $\{Z_{u,i}^{v}\}_{i=1}^{m}$ (yellow), 
where a zone $Z$ is a tuple of images and the corresponding camera poses $Z_{i} = \{(o_j,p_j)\}_{1}^{|Z_i|}$.
Given the seen zones, and the camera pose from an unseen zone $p_{u, i}^v$, we need to infer a feature encoding of
the unseen zone $Z_{u, i}^v$. To perform this task, we first extract visual features $x$ from each RGB-D frame $o$
in the video using pretrained CNNs (see Sec.~\ref{sec:smt}). These features are concatenated with the corresponding
pose $p$ and projected using an MLP $\mathcal{M}$ to obtain the image-level embedding. The target features for the
unseen zone $Z_{u, i}^v$ are obtained by randomly sampling an image within the zone and extracting its
camera pose and features:

\begin{equation} \label{eqn:target_feature}
    f_{u, i}^v = \mathcal{M}([x, p]),\textrm{where } (x, p) \sim Z_{u, i}^v.
\end{equation}

We use a transformer-based encoder-decoder model~\cite{vaswani2017attention} to perform this task. Our model consists
of an environment encoder and a zone decoder which infers the zone features (see Fig.~\ref{fig:approach}, bottom).
The environment encoder uses the image-level embeddings $\mathcal{M}([x, p])$ from the input zones and
performs multi-headed self-attention to generate the environment embeddings $\mathcal{E}$.

The zone decoder attends to $\mathcal{E}$ using the camera pose from the unseen zone $p_{u, i}^v$ and predicts the
zone features as follows:

\begin{equation}
    \hat{f}_{u, i} = \textrm{ZoneDecoder}(\mathcal{E}, p_{u, i}^v).
\end{equation}

We transform all poses in the input zones relative to $p_{u, i}^v$ before encoding, which provides the model an
egocentric view of the world. The environment encoder, zone decoder, and the projection function $\mathcal{M}$ are
jointly trained using noise-contrastive estimation~\cite{gutmann2010noise}. We use $\hat{f}_{u, i}$ as the anchor and
$f_{u, i}^{v}$ from Eqn.~\ref{eqn:target_feature} as the positive. We sample negatives from other unseen zones in the
same video and from all zones in other videos. The loss for the $i^{\textrm{th}}$ unseen zone in video $v$ is:

\begin{equation} \label{eqn:contrastive_loss}
\vspace*{-0.05in}\hspace{-0.1in}
\resizebox{0.44\textwidth}{!}{
    $\mathcal{L}_i^v = -\textrm{log}~\frac{
        \textrm{sim}(\hat{f}_{u,i}, f_{u,i}^v)
    }
    {
        \sum \limits_{j=1}^{m} \textrm{sim}(\hat{f}_{u,i}, f_{u,j}^v) +
        \sum \limits_{w \neq v, k} \textrm{sim}(\hat{f}_{u,i}, f_{k}^w)
    }$
}
\end{equation}
where $\textrm{sim}(q, k) = \textrm{exp}\big(\frac{q\cdot k}{|q||k|} \frac{1}{\tau}\big)$ and $\tau$ is a temperature
hyperparameter. The idea is to predict zone representations that are closer to the ground truth, while being
sufficiently different from the negative zones. Since the unseen zones have only limited overlap with the seen zones,
the model needs to effectively reason about the geometric and semantic context in the seen zones to differentiate the
positive from the negatives. We discourage the model from simply capturing video-specific textures and patterns by
sampling negatives from within the same video.

\begin{figure*}
    \centering
        \vspace*{-0.3in}
    \includegraphics[width=\textwidth,trim={0 8.5cm 3.5cm 0},clip]{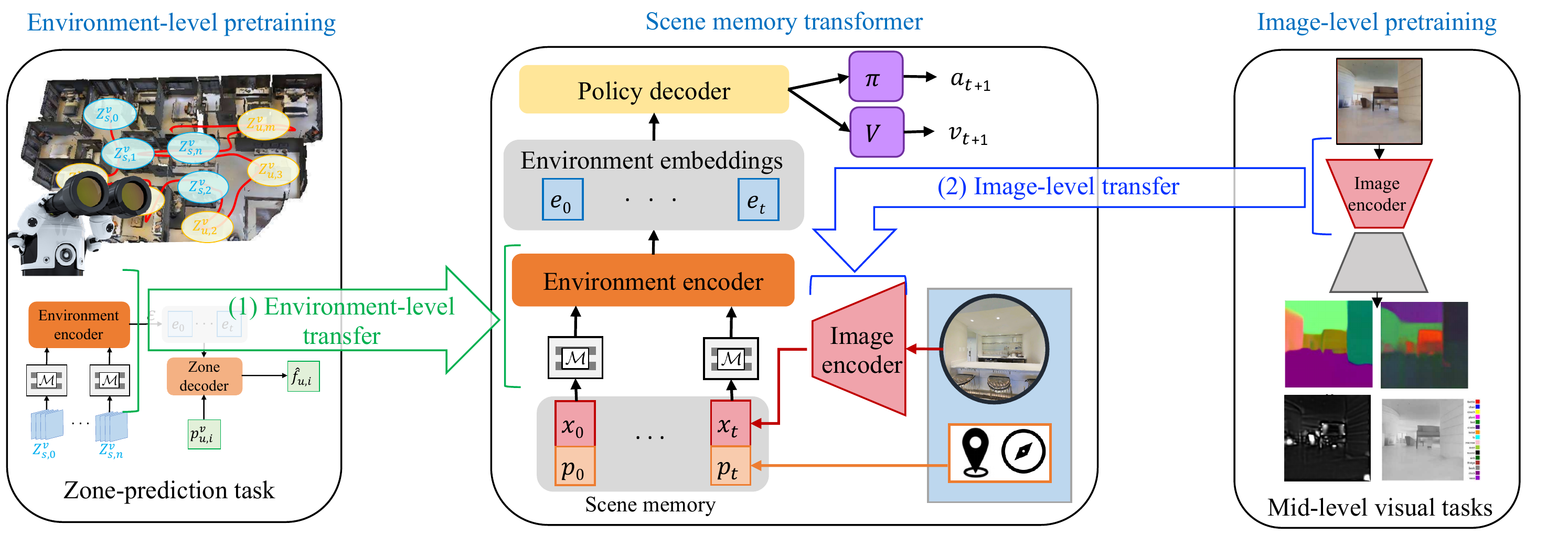}
    \vspace*{-0.2in}
    \caption{
        \small \textbf{Integrating environment-level pre-training for navigation:}
        \textbf{Left:} The first level of transfer occurs for the environment-level representations. We transfer the
        proposed EPC environment encoder and projection function $\mathcal{M}$ that are pre-trained for zone prediction.
        \textbf{Right:} The second level of transfer occurs for the image-level representations. We transfer a pre-trained
        MidLevel image encoder~\cite{sax2020learning} to generate visual features for each input in the scene memory.
        \textbf{Center:} To train the SMT on a task, we keep the visual features frozen, and finetune the environment
        encoder and projection function $\mathcal{M}$ with the rest of the SMT model.
    }
    \label{fig:smt}
\end{figure*}

\subsection{Environment embeddings for embodied agents} \label{sec:smt}

Having introduced our approach to learn environment embeddings in a self-supervised fashion, we now briefly overview how
these embeddings are used for agents performing navigation-oriented tasks. To this end, we integrate our pre-trained
environment encoder into the Scene Memory Transformer (SMT)~\cite{fang2019scene}. Our choice of SMT is motivated by the
recent successes of transformers in both NLP~\cite{devlin2018bert} and vision~\cite{sun2019videobert,fang2019scene}.
However, our idea is potentially applicable to other forms of memory models as well.

We briefly overview the SMT architecture (see Fig.~\ref{fig:smt}, center). It consists of a scene memory that stores
visual features $\{x_{i}\}_{i=0}^{t}$ and agent poses $\{p_{i}\}_{i=0}^{t}$ generated from the observations seen during
an episode. The environment encoder uses self-attention on the history of observations to generate a richer set of environment
embeddings $\{e_{i}\}_{i=1}^{t}$. At a given time-step $t+1$, the policy decoder attends to the environment embeddings using
the inputs $o_{t+1}$, which consist of the visual feature $x$ and agent pose $p$ at time $t+1$. The outputs of the policy
decoder are used to sample an action $a_{t+1}$ and estimate the value $v_{t+1}$. We detail each component in the Appendix.

To incorporate our EPC environment embeddings, we modify two key components from the original SMT model. First, and most
importantly, we initialize the environment encoder with our pre-trained EPC (see Fig.~\ref{fig:smt}, left). Second, we replace
the end-to-end trained image encoders with MidLevel features that are known to be useful across a variety of embodied
tasks~\cite{sax2020learning} (see Fig.~\ref{fig:smt}, right).\footnote{We pick MidLevel features~\cite{sax2020learning}
due to their demonstrated strong performance, though alternate image encoders are similarly applicable.}
We consider two visual modalities as inputs: RGB and depth.
For RGB, we extract features from the pre-trained models in the max-coverage set proposed by~\cite{sax2020learning}.
These include surface normals, keypoints, semantic segmentation, and 2.5D segmentation.  For depth, we extract features
from pre-trained models that predict surface normals and keypoints from depth~\cite{zamir2020robust}. For training the
model on a navigation task, we keep the visual features frozen, and only finetune the environment encoder, policy decoder,
policy $\pi$, and value function $V$.
\section{Experiments}\label{sec:results}

First, we review the experimental setup for the downstream navigation tasks (Sec.~\ref{sec:setupnav}). Next, we detail the
self-supervised learning setup and visualize the learned EPC embeddings (Sec.~\ref{sec:zoneacc}). We then evaluate the
pre-trained EPC environment embeddings on multiple downstream tasks that require an embodied agent to move intelligently
through an unmapped environment (Sec.~\ref{sec:navacc}). Finally, we evaluate the sensitivity of self-supervised learning
to noise in the video data (Sec.~\ref{sec:ablations}), and assess noise robustness of the learned policies on downstream
tasks (Sec.~\ref{sec:noise_robustness}).

\begin{figure*}
    \centering
    \includegraphics[width=1.0\textwidth,trim={0 0cm 0.0 0},clip]{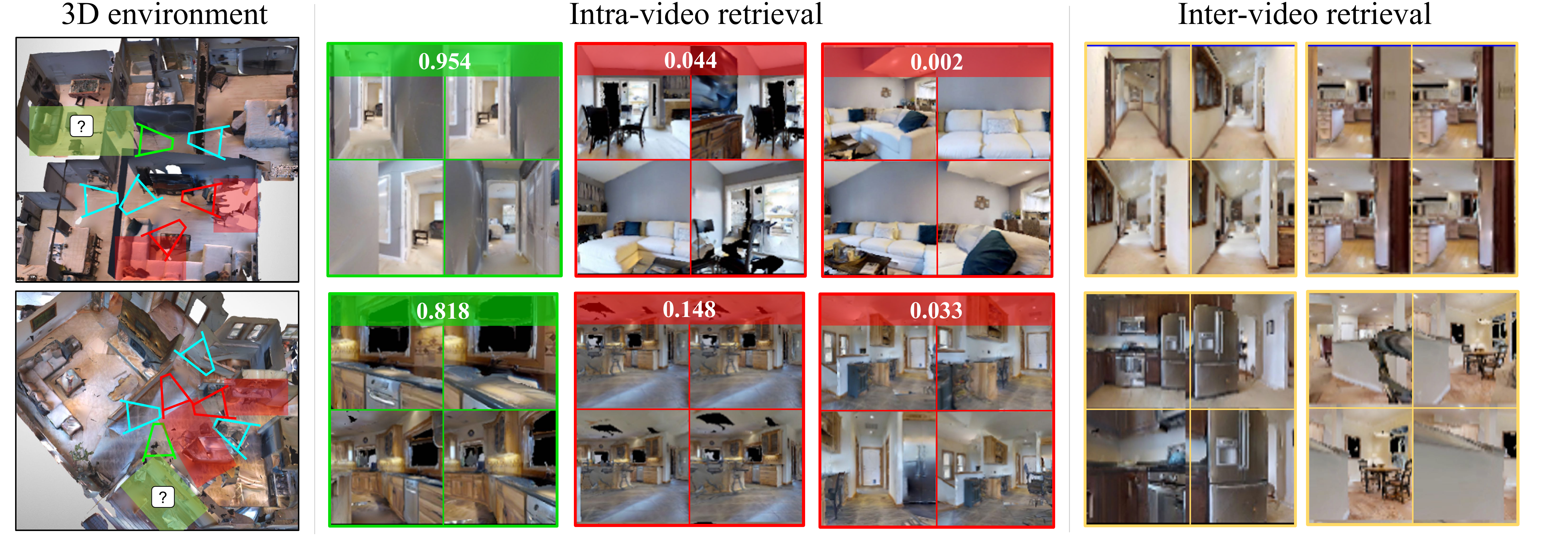}
    \caption{
        \small Each row shows one zone prediction example.
        \textbf{Left:} Top-down view of the 3D environment from which the video was sampled. The cyan viewing frusta correspond
        to the average pose for three input zones. Given the images and camera poses from each input zone, and a target camera
        pose (green frustum), the model predicts the corresponding zone feature (the masked green zone).
        \textbf{Center:} Given the inferred feature, we rank three masked (unobserved) zones from within the same video,
        where green is the positive zone and the red are the negatives. For each zone, we show four randomly sampled images
        along with the retrieval confidence. Our method  retrieves the positive with high confidence.  The model correctly
        predicts the existence of the narrow corridor (top row) and a kitchen counter (bottom row) given the target poses.
        \textbf{Right:} Top two retrieved zones from \emph{other} videos that are closest to the inferred feature. The features
        predicted by the model are general enough to retrieve related concepts from other videos (narrow corridors and kitchens).
    }
    \label{fig:zone_prediction_qual}
\end{figure*}

\subsection{Experimental setup for downstream navigation}\label{sec:setupnav}

We perform experiments on the Habitat simulator~\cite{savva2019habitat} with Matterport3D (MP3D)~\cite{chang2017matterport}
and Gibson~\cite{xia2018gibson}, two challenging and photorealistic 3D datasets with $\sim 90$ and $500$ scanned real-world
indoor environments, respectively. Our observation space consists of $171 \times 128$ RGB-D observations and odometry sensor
readings that provide the relative agent pose $p = (x, y, \theta)$ w.r.t the agent pose at $t=0$. Our action space consists
of: \textsc{move-forward} by $25\si{cm}$, \textsc{turn-left} by $30^{\circ}$, and \textsc{turn-right} by $30^{\circ}$. For all
methods, we assume noise-free actuation during training for simplicity. We evaluate with both noise-free and noisy sensing (pose, depth).

We use MP3D for interactive RL training, and reserve Gibson for evaluation. We use the default train/val/test split for MP3D~\cite{savva2019habitat}
for 1000-step episodes. For Gibson, which has smaller environments, we evaluate on the $14$ validation environments for 500-step
episodes.  Following prior work~\cite{ramakrishnan2020occant,chaplot2020learning}, we divide results on Gibson into small
and large environments.

We evaluate our approach on three standard tasks from the literature:

\noindent 1. \textbf{Area coverage}~\cite{chen2019learning,chaplot2020learning,ramakrishnan2020exploration}: The agent is rewarded
for maximizing the area covered (in $\si{m^2}$) within a fixed time budget. \vspace*{0.05in}\\
\noindent 2. \textbf{Flee}~\cite{gordon_iccv19}: The agent is rewarded for maximizing the flee distance (in $\si{m}$), i.e.,
the geodesic distance between its starting location and the terminal location, for fixed-length episodes. \vspace*{0.05in}\\
\noindent 3. \textbf{Object coverage}~\cite{fang2019scene,ramakrishnan2020exploration}: The agent is rewarded for maximizing
the number of categories of objects covered during exploration (see Appendix). Since Gibson lacks extensive object annotations,
we evaluate this task only on MP3D.

Together, these tasks capture different forms of geometric and semantic inference in 3D environments (e.g., area/object coverage
encourage finding large open spaces/new objects, respectively).
We compare our approach to the following baselines:\vspace*{0.05in}\\
\para{Scratch baselines} We randomly initialize the visual encoders and policy and train them end-to-end for each task. Images are
encoded using ResNet-18. Agent pose and past actions are encoded using FC layers. These are concatenated to obtain the features
at each time step. We use three temporal aggregation schemes. \emph{Reactive (scratch)} has no memory. \emph{RNN (scratch)} uses
a 2-layer LSTM as the temporal memory. \emph{SMT (scratch)} uses a Scene Memory Transformer for aggregating observations~\cite{fang2019scene}.\vspace*{0.05in}\\
\para{SMT (MidLevel)} extracts image features from pre-trained encoders that solve various mid-level perceptual tasks~\cite{sax2020learning}.
This is an ablation of our model from Sec.~\ref{sec:smt} that uses the same image features, but randomly initializes the environment encoder.
This SoTA image-level encoder is a critical baseline to show the impact of our proposed EPC environment-level encoder.\vspace*{0.05in}\\
\para{SMT (Video)} Inspired by Dense Predictive Coding~\cite{han2019video}, this baseline uses MidLevel features and pre-trains
the environment encoder as a video-level model using the same training videos as our model. It uses 25 consecutive frames as
inputs and predicts the features sampled from the next 15 frames (following timespans used in prior work~\cite{han2019video}).
We mask out the camera poses in the inputs and query based on the time (not pose). We train the model using the NCE loss
in Eqn.~\ref{eqn:contrastive_loss}.\vspace*{0.05in}\\
\para{OccupancyMemory} This is similar to the SoTA Active Neural SLAM model~\cite{chaplot2020learning} that maximizes
area coverage, but upgraded to use ground-truth depth to build the map (instead of RGB) and a state-of-the-art
PointNav agent~\cite{wijmans2019decentralized} for low-level navigation (instead of a planner). It represents the environment as a
top-down occupancy map.

All models are trained in PyTorch~\cite{paszke2019pytorch} with DD-PPO~\cite{wijmans2019decentralized} for 15M frames with
64 parallel processes and the Adam optimizer. See Appendix.

\begin{table*}[t!]
\begin{minipage}{\linewidth}
\centering
\scalebox{0.75}{
    \begin{tabular}{@{}ccccccccc@{}}
    \toprule
                       &   \multicolumn{3}{c}{Area coverage  (m$^2$)}       & \multicolumn{3}{c}{Flee     (m)}             &   \multicolumn{2}{c}{Object coverage (\#obj)}  \\
    \cmidrule(lr){2-4}\cmidrule(lr){5-7}\cmidrule(lr){8-9}
    Method                    &          Gibson-S     &      Gibson-L          &             MP3D       &           Gibson-S     &         Gibson-L       &          MP3D          &          MP3D-cat.     &          MP3D-inst.    \\ \midrule
    Reactive (scratch)        &     $ 17.4 \pm   0.2$ &      $ 22.8 \pm   0.6$ &      $ 68.0 \pm   1.3$ &      $  1.9 \pm   0.1$ &      $  2.5 \pm   0.3$ &      $  5.1 \pm   0.3$ &      $  6.2 \pm   0.0$ &      $ 19.0 \pm   0.2$ \\
    RNN (scratch)             &     $ 20.6 \pm   0.4$ &      $ 28.6 \pm   0.3$ &      $ 79.0 \pm   2.0$ &      $  2.3 \pm   0.2$ &      $  2.8 \pm   0.4$ &      $  5.8 \pm   0.0$ &      $  6.0 \pm   0.0$ &      $ 18.6 \pm   0.2$ \\
    SMT (scratch)             &     $ 23.0 \pm   0.6$ &      $ 32.3 \pm   0.8$ &      $104.8 \pm   2.2$ &      $  3.2 \pm   0.2$ &      $  4.4 \pm   0.4$ &      $  6.9 \pm   0.6$ &      $  7.0 \pm   0.2$ &      $ 23.2 \pm   0.9$ \\ \midrule
    SMT (MidLevel)            &     $ 29.1 \pm   0.1$ &      $ 47.2 \pm   1.6$ &      $155.6 \pm   2.0$ &      $  4.2 \pm   0.0$ &      $  6.0 \pm   0.4$ &      $ 10.6 \pm   0.3$ &      $  7.6 \pm   0.2$ &      $ 26.8 \pm   0.6$ \\
    SMT (Video)               &     $ 28.8 \pm   0.4$ &      $ 47.6 \pm   2.4$ &      $141.2 \pm   4.4$ &      $  4.0 \pm   0.0$ &      $  6.5 \pm   0.4$ &      $ 10.8 \pm   0.6$ &      $  7.5 \pm   0.1$ &      $ 26.0 \pm   0.2$ \\
    OccupancyMemory           &     $ 29.4 \pm   0.0$ & $ \bm{67.4} \pm   0.9$ &      $155.6 \pm   1.4$ &      $  2.8 \pm   0.0$ &      $  7.0 \pm   0.4$ & $ \bm{14.1} \pm   0.6$ &      $  7.8 \pm   0.1$ &      $ 27.8 \pm   0.4$ \\
    EPC                       & $\bm{31.5} \pm   0.1$ &      $ 62.2 \pm   1.0$ & $\bm{172.4} \pm   0.6$ & $  \bm{4.4} \pm   0.0$ & $  \bm{8.0} \pm   0.4$ &      $ 12.6 \pm   0.2$ & $  \bm{9.0} \pm   0.0$ & $ \bm{36.4} \pm   1.0$ \\
    \bottomrule
    \end{tabular}
}\vspace*{-0.05in}
\caption{
    \small\textbf{Downstream task performance} at the end of the episode. Gibson-S/L means small/large. MP3D-cat./inst. means
    categories/instances. All methods are evaluated on three random seeds.  Here EPC uses video walkthroughs collected
    by an exploration agent that maximizes its area coverage. See Appendix for performance vs.~time step plots.
}
\label{tab:nav_quant}
\end{minipage}
\vspace{-0.2cm}
\begin{minipage}{\linewidth}
    \centering
    \vspace*{0.1in}
    \includegraphics[width=\textwidth,trim={0 12.2cm 0.0cm 0}]{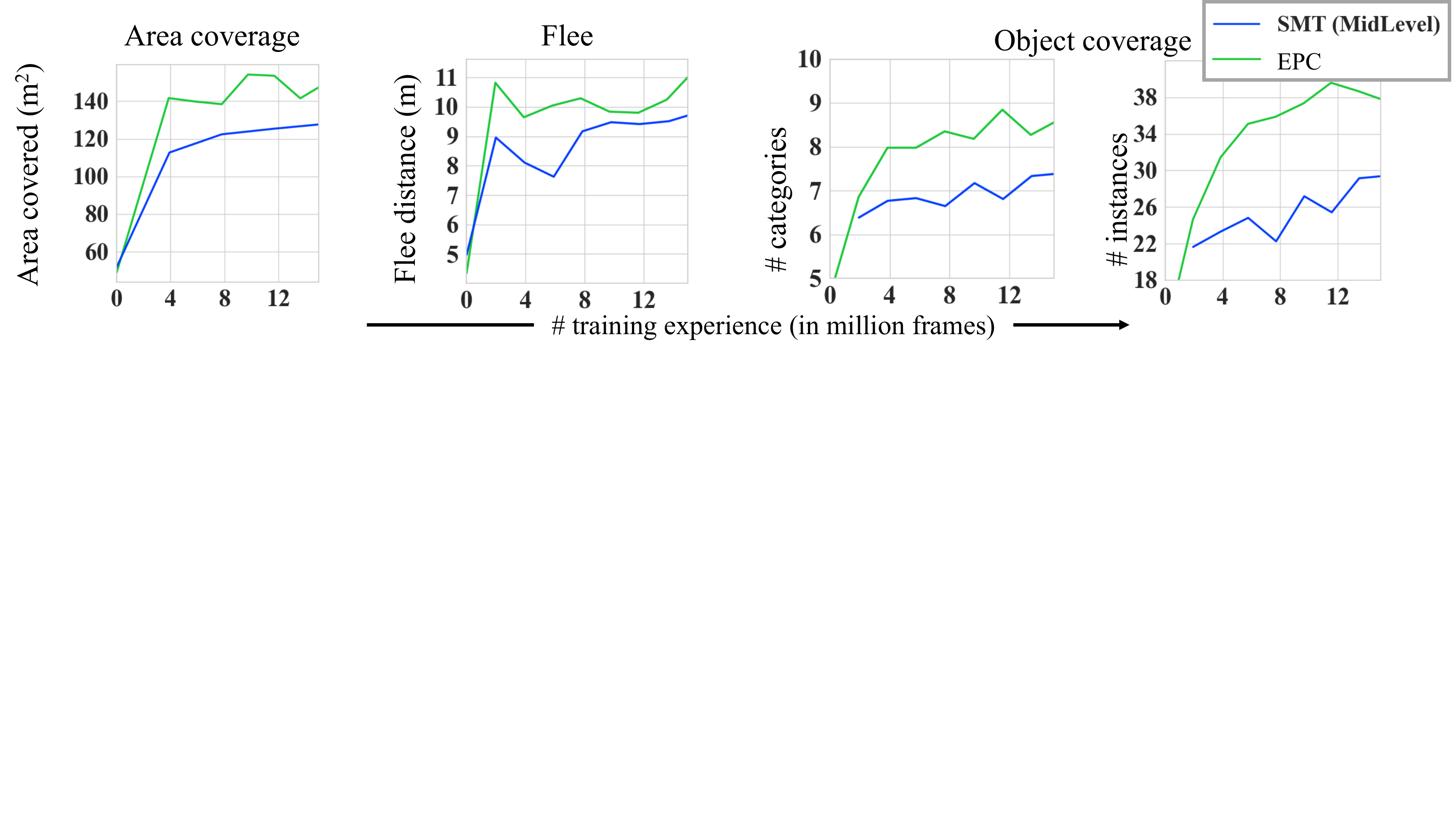}
    \vspace*{0.1cm}
    \captionof{figure}{\small
       \textbf{Sample efficiency} on Matterport3D val split. Our environment-level pre-training leads to 4-8$\times$ training
       sample efficiency when compared to SoTA image-level pre-training. See Appendix for Gibson plots.
    }
    \label{fig:nav_quant}
\end{minipage}
\end{table*}

\subsection{Self-supervised learning with EPC}\label{sec:zoneacc}

We generate walkthroughs for self-supervised learning from $332$ Gibson training environments. Note that these environments are
not accessible to the RL agent for interactive training. 

We collect the video data using two policies: 1) an SMT (scratch) agent that
was trained to perform area-coverage on MP3D, and 2) a heuristic navigation agent that moves forward until colliding, then turns
(cf.~Sec.~\ref{sec:ablations}). We test the impact of each video source separately below.

In both cases, the agents explore each Gibson environment starting from multiple locations and gather the RGB-D and odometer readings
for 500 steps per video. This results in $\sim$5,000 videos per agent, which we divide into an 80-20 train/val split. We use these videos to
pre-train environment encoders on the EPC zone prediction task for 50 epochs. The hyperparameters are provided in Appendix~\ref{appsec:hyperparameters}.
We qualitatively analyze the masked zone prediction results from EPC in Fig.~\ref{fig:zone_prediction_qual}.

\subsection{Downstream task performance}\label{sec:navacc}

Now we transfer these features to downstream navigation tasks. Tab.~\ref{tab:nav_quant} shows the results. On both datasets,
we observe the following ordering:

\begin{equation}
\textrm{Reactive (scratch)} ~<~\textrm{RNN (scratch)}~<~\textrm{SMT (scratch)}.
\end{equation}
This is in line with results reported by~\cite{fang2019scene} and verifies our implementation of SMT.  Using MidLevel features for
SMT leads to significant gains in performance versus training image encoders from scratch.

Environment Predictive Coding (EPC) effectively combines environment-level and image-level pretraining to provide substantial
improvements compared to only image-level pretraining from SMT (MidLevel), particularly for larger environments.
Furthermore, SMT (Video)---the video-level pre-training strategy---significantly
underperforms EPC, as it ignores the underlying spatial information during self-supervised learning. This highlights EPC's value
in representing the underlying 3D spaces of the walkthroughs instead of treating them simply as video frames. EPC
also competes closely and even slightly outperforms the state-of-the-art OccupancyMemory on the geometric tasks
area coverage and flee, while providing a significant gain on object coverage. Thus, our model competes
strongly with \emph{the purely geometric} representation model on the tasks that the latter was designed for, while
outperforming it significantly on the semantic task. Note that OccupancyMemory performs poorly on Flee
in Gibson-S since its goal sampling strategy is biased towards larger environments. See Appendix~\ref{appsec:occmem}.

Finally, in Fig.~\ref{fig:nav_quant}, we can see that environment-level pre-training from EPC offers significantly higher
sample efficiency: our method reaches the best performance of SMT (MidLevel) 4-8$\times$ faster. This advantage persists
even after accounting for the $\sim$2M frames of off-policy experience in the video data (see Tab.~\ref{tab:offpolicy}).
This confirms our hypothesis: transferring environment-level representations learned via contextual reasoning can help embodied
agents learn faster compared to the current approach of transferring image-level encoders alone.

\begin{table*}[t]
\centering
\scalebox{0.80}{
    \begin{tabular}{@{}lcccccccc@{}}
    \toprule
                                  &              \multicolumn{3}{c}{Area coverage  (m$^2$)}             &                  \multicolumn{3}{c}{Flee     (m)}              &    \multicolumn{2}{c}{Object coverage (\#obj)}   \\ \cmidrule(lr){2-4}\cmidrule(lr){5-7}\cmidrule(lr){8-9}
    Method                        &         Gibson-S      &      Gibson-L     &            MP3D         &         Gibson-S      &        Gibson-L    &       MP3D        &        MP3D-cat.      &     MP3D-inst.           \\
    SMT (MidLevel)                &     $ 29.1  \pm  0.1$ & $ 47.2 \pm  1.7 $ &     $155.7  \pm  2.0  $ &    $  4.2  \pm  0.0 $ & $ 6.0   \pm  0.4 $ & $ 10.6 \pm  0.3 $ &    $ 7.6   \pm  0.1 $ &  $ 26.8  \pm  0.6  $     \\
    EPC                           &     $ 31.5  \pm  0.1$ & $ 62.2 \pm  1.0 $ &     $172.5  \pm  0.6  $ &    $  4.5  \pm  0.0 $ & $ 8.0   \pm  0.5 $ & $ 12.7 \pm  0.2 $ & $\bm{9.0}  \pm  0.1 $ &  $ 36.4  \pm  1.0  $     \\
    EPC w/ noisy depth            &     $ 31.4  \pm  0.4$ & $ 60.5 \pm  1.4 $ &     $171.3  \pm  5.0  $ &    $  4.2  \pm  0.1 $ & $ 7.6   \pm  1.0 $ & $ 11.5 \pm  0.5 $ &    $ 8.6   \pm  0.1 $ &  $ 35.2  \pm  0.6  $     \\
    EPC w/ noisy depth and pose   &  $\bm{32.2} \pm  0.1$ & $ 63.6 \pm  3.2 $ & $\bm{181.0} \pm  1.5  $ &  $\bm{4.9} \pm  0.2 $ & $ 7.9   \pm  0.2 $ & $ 13.3 \pm  0.4 $ &    $ 8.5   \pm  0.2 $ &  $ 33.8  \pm  1.2  $     \\
    EPC w/ heuristic video policy &     $ 31.4  \pm  0.3$ & $ 61.6 \pm  2.3 $ &     $176.4  \pm  4.8  $ &    $  4.5  \pm  0.2 $ & $ 7.1   \pm  0.4 $ & $ 12.3 \pm  0.4 $ &    $ 8.5   \pm  0.1 $ &  $ 35.9  \pm  0.1  $     \\
    \bottomrule
    \end{tabular}
}
\caption{
    Impact of noisy video data (corrupted pose and/or depth) and a simple heuristic policy for video generation on EPC self-supervised learning.
    EPC maintains its advantage over the SMT (MidLevel) which randomly initializing the environment encoders.
}
\label{tab:sensitivity_ssl}
\end{table*}

\begin{table*}[t]
\centering
\scalebox{0.8}{
    \begin{tabular}{@{}lccccccccc@{}}
    \toprule
                              &\multicolumn{3}{c}{Area coverage  (m$^2$)}      &                  \multicolumn{3}{c}{Flee     (m)}               &           \multicolumn{3}{c}{Object cov. (\#cat.)}           \\
   \cmidrule(lr){2-4}\cmidrule(lr){5-7}\cmidrule(lr){8-10}
    Method                    &           NF         &          N-D         &         N-D,P           &     NF              &           N-D       &        N-D,P        &          NF        &           N-D      &        N-D,P       \\
    SMT (MidLevel)            &     $155.7  \pm 2.0$ &     $ 145.1 \pm 2.3$ &     $ 134.2 \pm   1.8$  &     $10.6  \pm 0.3$ &     $10.6  \pm 0.6$ &     $10.8  \pm 0.4$ &     $7.6  \pm 0.2$ &     $7.3  \pm 0.1$ &     $7.3  \pm 0.2$ \\
    OccupancyMemory           &     $155.6  \pm 1.4$ &     $  86.6 \pm 2.2$ &     $  85.2 \pm   2.4$  & $\bm{14.1} \pm 0.6$ &     $10.9  \pm 0.2$ &     $10.2  \pm 0.3$ &     $7.8  \pm 0.1$ &     $5.8  \pm 0.0$ &     $5.8  \pm 0.0$ \\
    EPC                       & $\bm{172.5} \pm 0.6$ &  $\bm{161.6}\pm 3.1$ &  $\bm{159.3}\pm   2.0$  &     $12.7  \pm 0.2$ & $\bm{12.0} \pm 0.8$ & $\bm{12.0} \pm 0.1$ & $\bm{9.0} \pm 0.1$ & $\bm{8.5} \pm 0.1$ & $\bm{8.5} \pm 0.3$ \\
    \bottomrule
    \end{tabular}
}
\caption{
    Comparing robustness to sensor noise on downstream tasks in Matterport3D. Note: NF denotes noise free sensing, N-D denotes
    noisy depth (and noise-free pose), and N-D,P denotes noisy depth and pose. See Appendix~\ref{sec:appendix_noise_robustness} for full results.
}
\label{tab:noise_robustness}
\end{table*}

\subsection{Sensitivity analysis of self-supervised learning}\label{sec:ablations}

Next we analyze the sensitivity of EPC to 1) sensory noise in the videos and 2) the exploration strategy used for video data collection.
Specifically, we inject noise in the depth and pose data from the walkthrough videos using existing noise models
from~\cite{choi2015robust} and~\cite{ramakrishnan2020occant}. The depth noise model combines disparity-based quantization,
high-frequency noise, and low-frequency distortion~\cite{choi2015robust}. The odometry noise is based on data collected from
a LoCoBot robot~\cite{ramakrishnan2020occant,chaplot2020learning}. We also replace the video walkthroughs from the area-coverage
agent with an equivalent amount of data collected by a simple heuristic used in prior work~\cite{chen2019learning,ramakrishnan2020exploration}.
The heuristic instructs the video agent to move as follows: move forward until colliding, then turn left or right by a random amount, then continue moving forward. 

Tab.~\ref{tab:sensitivity_ssl} shows the impact of each of these changes on the downstream task performance. For reference,
we compare these with random initilization of the environment encoder in SMT (MidLevel). Our approach EPC is reasonably robust to changes
in the video data during SSL training. The performance remains stable when noise is injected into depth inputs. While it starts to decline on object coverage when we further inject
noise into the pose inputs, EPC still retains its advantages over SMT (MidLevel). Note that we do not employ any noise-correction mechanisms,
which could better limit this decline~\cite{chaplot2020learning,ramakrishnan2020occant}. Finally, the performance is not significantly impacted
when we use video data from the simple exploration heuristic, emphasizing that EPC does not require a strong exploration policy for the agent
that generates the self-supervised training videos, nor does it require a tight similarity between the tasks demonstrated in the videos and the downstream tasks.

\begin{table*}[t!]
    \begin{minipage}{\linewidth}
    \centering
    \scalebox{0.85}{
        \begin{tabular}{@{}ccccccccc@{}}
        \toprule
                                  &             \multicolumn{3}{c}{Area coverage  (m$^2$)}                  &                     \multicolumn{3}{c}{Flee (m)}                         &     \multicolumn{2}{c}{Object coverage (\#obj)}  \\
        \cmidrule(lr){2-4}\cmidrule(lr){5-7}\cmidrule(lr){8-9}
        Method                    &          Gibson-S     &      Gibson-L          &             MP3D       &           Gibson-S     &         Gibson-L       &          MP3D          &          MP3D-cat.     &          MP3D-inst.     \\ \midrule
        SMT (Video)  @ 15M        &     $ 28.8 \pm   0.4$ &      $ 47.6 \pm   2.4$ &      $141.2 \pm   4.4$ &      $ 4.0  \pm   0.0$ &      $  6.5 \pm   0.4$ &      $ 10.8 \pm   0.6$ &      $  7.5 \pm   0.1$ &      $ 26.0 \pm   0.2$  \\
        SMT (Video)  @ 13M        &     $ 28.4 \pm   0.3$ &      $ 44.7 \pm   2.5$ &      $126.2 \pm   3.4$ &      $ 3.7  \pm   0.1$ &      $  4.9 \pm   0.2$ &      $ 10.0 \pm   0.5$ &      $  7.2 \pm   0.1$ &      $ 24.3 \pm   0.6$  \\
        EPC  @ 15M                & $\bm{31.5} \pm   0.1$ &      $ 62.2 \pm   1.0$ &      $172.4 \pm   0.6$ &      $ 4.4  \pm   0.0$ & $  \bm{8.0} \pm   0.4$ &      $ 12.6 \pm   0.2$ & $  \bm{9.0} \pm   0.0$ & $ \bm{36.4} \pm   1.0$  \\
        EPC  @ 13M                &     $ 31.2 \pm   0.2$ &   $\bm{64.0}\pm   0.5$ &  $\bm{176.3}\pm   3.8$ &  $ \bm{4.6} \pm   0.2$ &      $  7.1 \pm   0.8$ &  $ \bm{13.8}\pm   0.7$ & $  \bm{9.0} \pm   0.0$ & $ \bm{36.4} \pm   1.0$  \\

        \bottomrule
        \end{tabular}
    }\vspace*{-0.05in}
    \caption{
            \small\textbf{Offsetting on-policy interaction with video-level experience:} Each walkthrough video contains $\sim$500 frames and the dataset contains $\sim$2M frames.
            While this is \emph{off-policy walkthrough data} used only for representation learning, we nonetheless re-evaluate the checkpoints of ``EPC" and ``SMT (Video)" before 15M-2M = 13M frames
            to account for the additional observations from the video data. The performance at 13M and 15M frames remains similar for EPC, whereas it decreases at 13M frames for SMT (Video).
            Thus, EPC maintains its advantages over the best baselines from Tab.~\ref{tab:nav_quant} after accounting for any additional experience from the video walkthroughs.
    }
    \label{tab:offpolicy}
    \end{minipage}
\end{table*}

\subsection{Robustness of learned policies to sensor noise}\label{sec:noise_robustness}

In the previous experiments, we assume the availability of ground-truth depth and pose sensors for downstream tasks (Tab.~\ref{tab:sensitivity_ssl} added pose
and depth noise to the walkthrough videos only).  Now, we relax these assumptions and re-evaluate all methods by injecting noise in the depth and pose sensors for downstream tasks (same noise
models from prior work that we applied in Sec.~\ref{sec:ablations}), without any noise-correction.  This is a common evaluation protocol for assessing noise
robustness~\cite{chen2019learning, ramakrishnan2020exploration}. We compare the top three methods on MP3D in Tab.~\ref{tab:noise_robustness}
and provide the complete set of results in Appendix~\ref{sec:appendix_noise_robustness}. As expected, the performance declines
slightly as we add noise to more sensors (depth, then pose). However, most approaches are reasonably stable. EPC outperforms
all methods when all noise sources are added. OccupancyMemory declines rapidly in the absence of noise-correction due to accumulated
errors in the map.   
\vspace*{-0.05in}
\section{Conclusions}
\vspace*{-0.05in}

We introduced Environment Predictive Coding, a self-supervised approach to learn environment-level representations
for embodied agents. By training on video walkthroughs generated by other agents,  our model learns to infer missing
content through a zone-prediction task. When transferred to multiple downstream embodied agent tasks, the resulting
embeddings lead to better performance and sample efficiency  compared to the current practice of transferring only
image-level representations. In future work, we plan to extend our idea for goal-driven tasks like PointNav and ObjectNav.

{\small
\bibliographystyle{ieee_fullname}
\bibliography{egbib}
}

\appendix

\vspace*{0.2in}
\textbf{\Large Appendix}

\section{Zone generation}

As discussed in the main paper, we generate zones by first clustering frames in the video based on their geometric overlap.
Here, we provide details on how this overlap is estimated. First, we project pixels in the image to 3D point-clouds using
the camera intrinsics and the agent pose. Let $D_i$, $p_i$ be the depth map and agent pose for frame $i$ in the video.
The agent's pose in frame $i$ can be expressed as $p_{i} = (\bm{R}_{i}, \bm{t}_{i})$, with $\bm{R}_{i}, \bm{t}_{i}$ representing
the \textit{agent's} camera rotation and translation in the world coordinates. Let $\bm{K} \in \mathbb{R}^{3\times 3}$ be the
intrinsic camera matrix, which is assumed to be provided for each video. We then project each pixel $x_{ij}$ in the depth map
$D_{i}$ to the 3D point cloud as follows:

\begin{equation}
    w_{ij} = \begin{bmatrix}
                \bm{R}_{i} & \bm{t}_{i}\\
                \bm{0} & 1
             \end{bmatrix} \bm{K}^{-1} x_{ij},~\forall j \in \{1, ..., S_i\}
\end{equation}

\noindent where $S_i$ is the total number of pixels in $D_{i}$. By doing this operation for each pixel, we can obtain the
point-cloud $\bm{W}_i$ corresponding to the depth map $D_i$. To compute the geometric overlap between two frames $i$ and $j$,
we estimate the overlap in their point-clouds $\bm{W}_i$ and $\bm{W}_j$. Specifically, for each point $w_i \in \bm{W}_i$,
we retrieve the nearest neighbor from $w_j \in \bm{W}_j$ and check whether the pairwise distance in 3D space is within a
threshold $\tau$: $|| w_i - w_j ||_2 < \tau$ . If this condition is satisfied, then a match exists for $w_i$.
Then, we define the overlap fraction $\psi(D_i, D_j)$ the fraction of points in $\bm{W}_i$ which have a match in $\bm{W}_{j}$.
This overlap fraction is computed pairwise between all frames in the video, and hierarchical agglomerative clustering is
performed using this similarity measure.

\section{Task details}

For the object coverage task, to determine if an object is covered, we check if it is within $3\si{m}$ of the agent, present
in the agent's field of view, and if it is not occluded~\cite{ramakrishnan2020exploration}. We use a shaped reward function:

\begin{equation}
R_{t} = O_{t}-O_{t-1} + 0.02 (C_{t} - C_{t-1}),
\end{equation}
where $O_{t}$, $C_{t}$ are the number of object categories and 2D grid-cells visited by time $t$ (similar to ~\cite{fang2019scene}).

\section{Scene memory transformer}

We provide more details about individual components of the Scene Memory Transformer~\cite{fang2019scene}. As discussed in the
main paper, the SMT model consists of a scene memory for storing the visual features $\{x_{i}\}_{i=0}^{t}$ and agent poses
$\{p_{i}\}_{i=0}^{t}$ seen during an episode. The environment encoder uses self-attention on the scene memory to generate a
richer set of environment embeddings $\{e_{i}\}_{i=1}^{t}$. The policy decoder attends to the environment embeddings using
the inputs $o_{t+1}$, which consist of the visual feature $x$, and agent pose $p$ at time $t+1$. The outputs of the policy
decoder are used to sample an action $a_{t+1}$ and estimate the value $v_{t+1}$. Next, we discuss the details of the
individual components.

\paragraph{\textsc{Scene memory}}

It stores the visual features derived from the input images and the agent poses at each time-step. Motivated by the ideas
from~\cite{sax2020learning}, we use mid-level features derived from various pre-trained CNNs for each input modality.
In this work, we consider two input modalities: RGB, and depth. For RGB inputs, we extract features from the pre-trained
models in the max-coverage set proposed in~\cite{sax2020learning}. These include surface normals, keypoints, semantic segmentation,
and 2.5D segmentation. For depth inputs, we extract features from pre-trained models that predict surface normals and keypoints
from depth~\cite{zamir2020robust}. For simplicity, we assume that the ground-truth pose is available to the agent in the form
of $(x_{t}, y_{t}, z_{t}, \theta_{t})$ at each time-step, where $\theta_{t}$ is the agent heading. While this can be relaxed by
following ideas from state-of-the-art approaches to Neural SLAM~\cite{chaplot2020learning,ramakrishnan2020occant}, we reserve
this for future work as it is orthogonal to our primary contributions.

\paragraph{\textsc{Attention mechanism}}

Following the notations from~\cite{vaswani2017attention}, we define the attention mechanism used in the environment encoder
and policy decoder. Given two inputs $X \in \mathbb{R}^{n_1 \times d_x}$ and $Y \in \mathbb{R}^{n_2 \times d_y}$, the attention
mechanism attends to $Y$ using $X$ as follows:

\begin{equation}
    \textrm{Attn}(X, Y) = \textrm{softmax}\bigg(\frac{Q_X K_Y^T}{\sqrt{d_k}}\bigg) V_Y
\end{equation}

where $Q_X \in \mathbb{R}^{n_1 \times d_k}, K_Y \in \mathbb{R}^{n_2 \times d_k}, V_Y \in \mathbb{R}^{n_2 \times d_v}$ are
the queries, keys, and values computed from $X$ and $Y$ as follows:  $Q_X = X W^q$, $K_Y = Y W^k$, and
$V_Y = Y W^v$. $W^q, W^k, W^v$ are learned weight matrices. The multi-headed version of Attn generates multiple sets of
queries, keys, and values to obtain the attended context $C \in \mathbb{R}^{n_1 \times d_v}$.

\begin{equation}
    \textrm{MHAttn}(X, Y) = \textrm{FC}([\textrm{Attn}^{h}(X, Y)]_{h=1}^{H}).
\end{equation}

We use the transformer implementation from PyTorch~\cite{paszke2019pytorch}. Here, the multi-headed attention block builds
on top of MHAttn by using residual connections, LayerNorm (LN) and fully connected (FC) layers to further encode the inputs.

\begin{equation}
\begin{gathered}
    \textrm{MHAttnBlock}(X, Y) = \textrm{LN}(\textrm{MLP}(H) + H)
\end{gathered}
\end{equation}

where $H = \textrm{LN}(\textrm{MHAttn}(X, Y) + X)$, and MLP has 2 FC layers with ReLU activations. The environment encoder
performs self-attention between the features stored in the scene memory to obtain the environment encoding $E$.

\begin{equation}
    E = \textrm{EnvironmentEncoder}(M) = \textrm{MHAttnBlock}(M, M).
\end{equation}

The policy decoder attends to the environment encodings $E$ using the current observation $x_t, p_t$.

\begin{equation}
    \textrm{PolicyDecoder}([x_t, p_t], E) = \textrm{MHAttnBlock}(\textrm{FC}([x_t, p_t]), E)
\end{equation}

We transform the pose vectors $\{p_{i}\}_{i=1}^{n}$ from the scene memory relative to the current agent pose $p_t$ as this
allows the agent to maintain an egocentric view of past inputs~\cite{fang2019scene}.

\section{Hyperparameters}\label{appsec:hyperparameters}

We detail the list of hyperparameter choices for different tasks and models in Tab.~\ref{tab:hyperparameters}.
For SMT (Video), we randomly sample 40 consecutive frames in the video and predict the final 15 frames from the initial
25 frames (based on Dense Predictive Coding~\cite{han2019video}). For EPC, we randomly mask out 4 zones in the video
and predict them from the remaining video. The hyperparameters are selected based on validation performance on the
downstream tasks.

\begin{table*}[h!]
\centering
\begin{tabular}{lc}
\toprule
\multicolumn{2}{c}{RL Optimization}                                             \\ \midrule
\multicolumn{1}{l}{Optimizer}                             & Adam                \\
\multicolumn{1}{l}{Learning rate}                         & 0.00025 - 0.001     \\
\multicolumn{1}{l}{$\#$ parallel actors}                  & 64                  \\
\multicolumn{1}{l}{PPO mini-batches}                      & 2                   \\
\multicolumn{1}{l}{PPO epochs}                            & 2                   \\
\multicolumn{1}{l}{PPO clip param}                        & 0.2                 \\
\multicolumn{1}{l}{Value loss coefficient}                & 0.5                 \\
\multicolumn{1}{l}{Entropy coefficient}                   & 0.01                \\
\multicolumn{1}{l}{Advantage estimation}                  & GAE                 \\
\multicolumn{1}{l}{Normalized advantage?}                 & Yes                 \\
\multicolumn{1}{l}{Training episode length}               & 1000                \\
\multicolumn{1}{l}{GRU history length}                    & 128                 \\
\multicolumn{1}{l}{$\#$ training steps (in millions)}     & 15                  \\ \midrule
\multicolumn{2}{c}{RNN hyperparameters}                                         \\ \midrule
\multicolumn{1}{l}{Hidden size}                           & 128                 \\
\multicolumn{1}{l}{RNN type}                              & LSTM                \\
\multicolumn{1}{l}{Num recurrent layers}                  & 2                   \\ \midrule
\multicolumn{2}{c}{SMT hyperparameters}                                         \\ \midrule
\multicolumn{1}{l}{Hidden size}                           & 128                 \\
\multicolumn{1}{l}{Scene memory length}                   & 500                 \\
\multicolumn{1}{l}{$\#$ attention heads}                  & 8                   \\
\multicolumn{1}{l}{$\#$ encoder layers}                   & 1                   \\
\multicolumn{1}{l}{$\#$ decoder layers}                   & 1                   \\ \midrule
\multicolumn{2}{c}{Occupancy memory hyperparameters}                            \\ \midrule
\multicolumn{1}{l}{Action space range}                    & $48\si{m} \times 48\si{m}$ \\
\multicolumn{1}{l}{$\#$ global action sampling interval}  & $25$                \\ \midrule
\multicolumn{2}{c}{Reward scaling factors for different tasks}                  \\
\multicolumn{1}{l}{Task}                                  & Reward scale        \\ \midrule
\multicolumn{1}{l}{Area coverage}                         & 0.3                 \\
\multicolumn{1}{l}{Flee}                                  & 1.0                 \\
\multicolumn{1}{l}{Object coverage}                       & 1.0                 \\ \midrule
\multicolumn{2}{c}{Self-supervised learning optimization}                       \\ \midrule
\multicolumn{1}{l}{Optimizer}                             & Adam                \\
\multicolumn{1}{l}{Learning rate}                         & 0.0001              \\
\multicolumn{1}{l}{Video batch size}                      & 20                  \\
\multicolumn{1}{l}{Temperature ($\tau$)}                    & 0.1                 \\
\bottomrule
\end{tabular}
\caption{\small Hyperparameters for training our RL and self-supervised learning models.}
\label{tab:hyperparameters}
\end{table*}

\section{Downstream task performance vs. time}

We show the downstream task performance as a function of time in Fig.~\ref{fig:nav_quant_supp}. We evaluate each model
with 3 different random seeds and report the mean and the $95\%$ confidence interval in the plots.

\begin{table*}[t]
\begin{minipage}{\linewidth}
    \centering
    \vspace*{0.1in}
    \includegraphics[width=\textwidth]{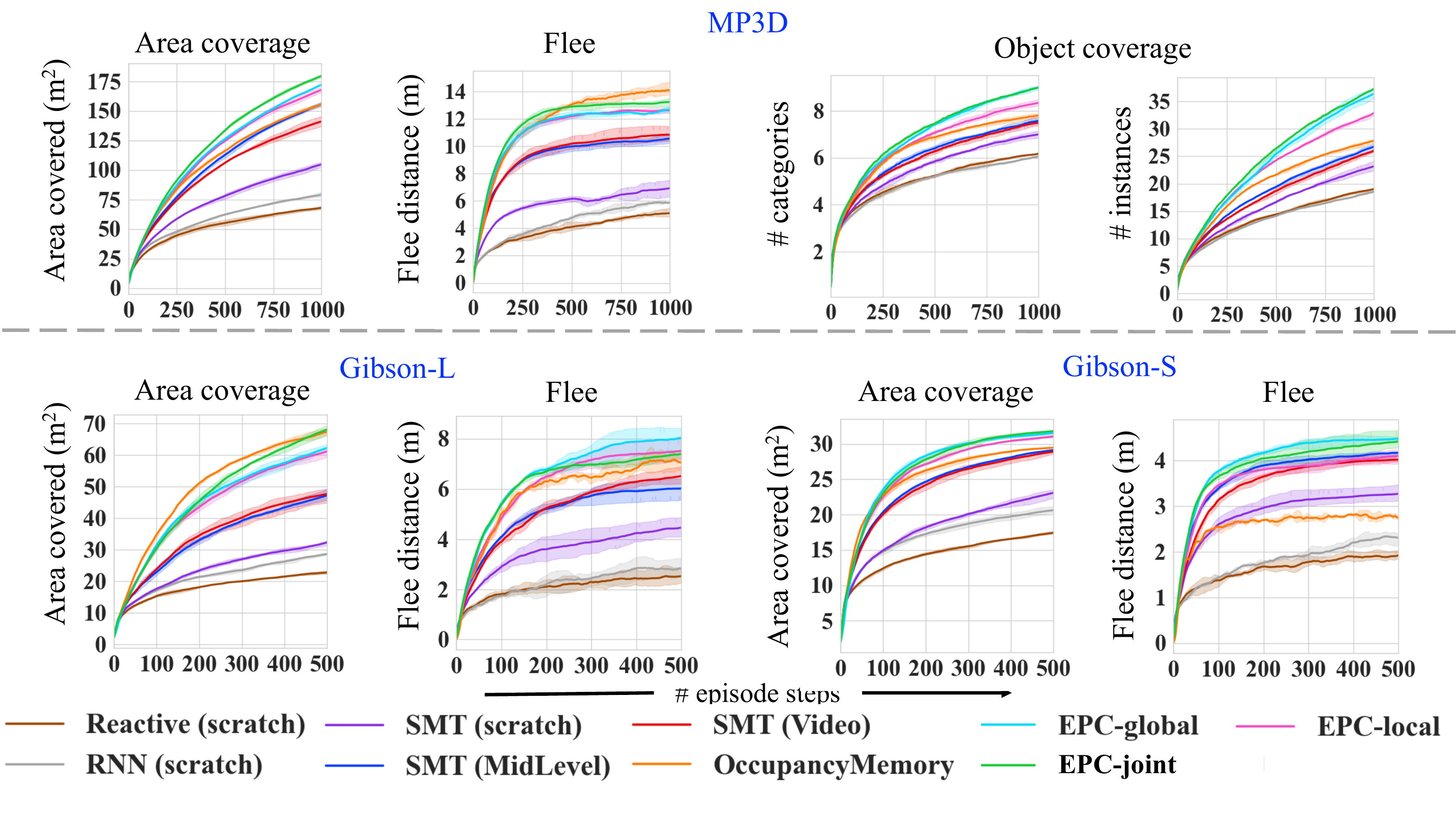}
    \captionof{figure}{\small
        We highlight the downstream task performance as a function of episode time on both Matterport3D and Gibson.
    }
    \label{fig:nav_quant_supp}
\end{minipage}
\end{table*}

\section{Sample efficiency curves on Gibson}
We plot the Gibson validation performance as a function of training experience in Fig.~\ref{fig:nav_sample_efficiency_supp}.
EPC achieves better sample efficiency through environment-level pre-training when compared to the image-level pre-training
baseline SMT (MidLevel).

\begin{table*}[t!]
\begin{minipage}{\linewidth}
    \centering
    \vspace*{0.1in}
    \includegraphics[width=0.6\textwidth,trim={0 10.5cm 14.0cm 0}]{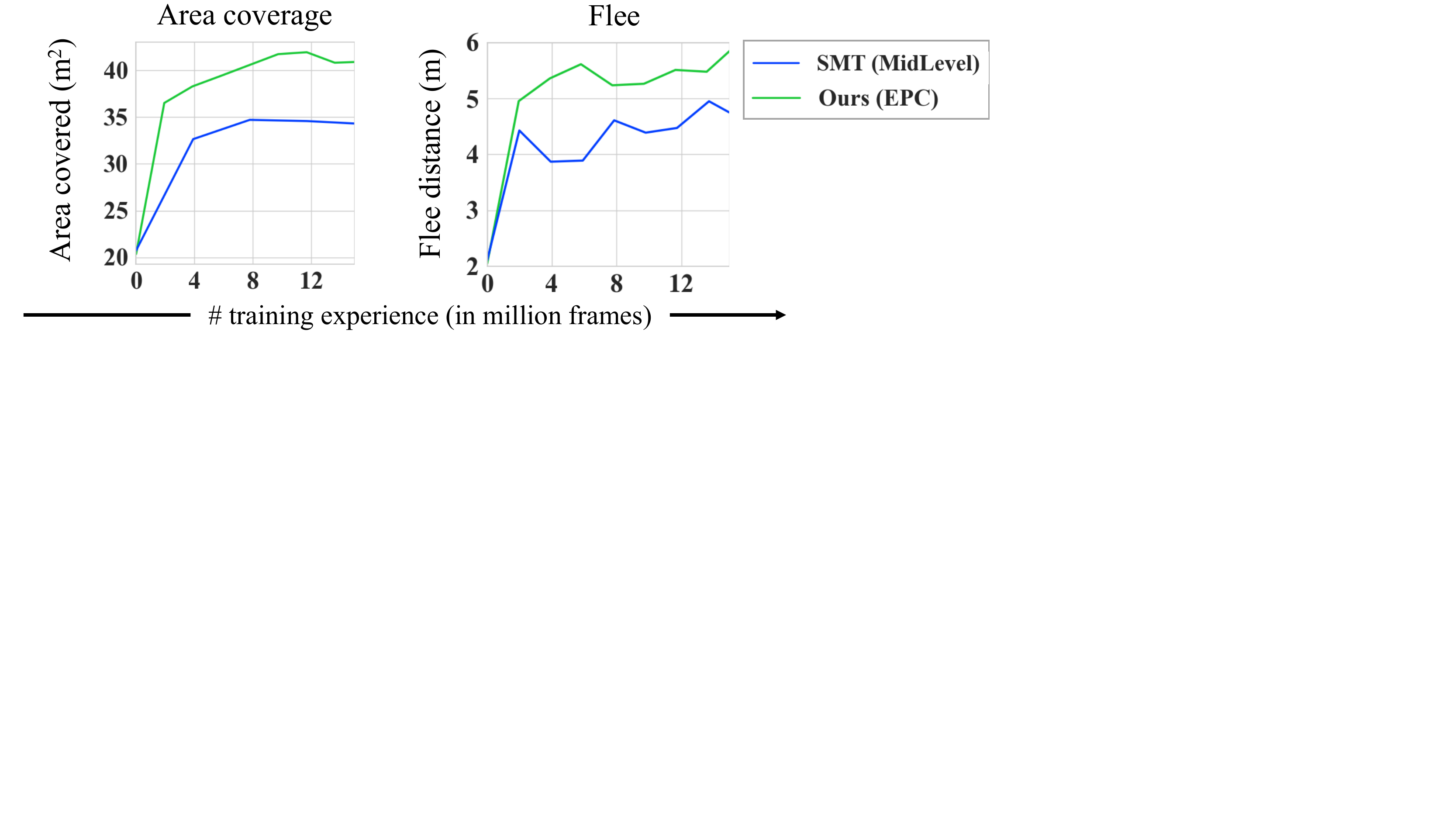}
    \captionof{figure}{\small
       \textbf{Sample efficiency} on Gibson val split. Our environment-level pre-training leads to 4-8$\times$ training
       sample efficiency when compared to SoTA image-level pre-training.
    }
    \label{fig:nav_sample_efficiency_supp}
\end{minipage}
\end{table*}

\section{Complete analysis of noise robustness in downstream tasks}
\label{sec:appendix_noise_robustness}

In Tab.~\ref{tab:noise_robustness} from the main paper, we compared the noise robustness of top three approaches on MP3D. Here,
we present the complete set of results for all methods on Gibson and MP3D in Tab.~\ref{tab:appendix_noise_robustness}.

\begin{table*}[t]
\centering
\scalebox{0.85}{
    \begin{tabular}{@{}lccccccccc@{}}
    \toprule
                              &                                                                                  \multicolumn{9}{c}{\textbf{Matterport3D}}                                                                             \\
                              &\multicolumn{3}{c}{Area coverage  (m$^2$)}                             &                  \multicolumn{3}{c}{Flee     (m)}               &           \multicolumn{3}{c}{Object cov. (\#cat.)}           \\
   \cmidrule(lr){2-4}\cmidrule(lr){5-7}\cmidrule(lr){8-10}
    Method                    &            NF        &           N-D        &           N-D,P         &         NF          &           N-D       &        N-D,P        &          NF        &           N-D      &        N-D,P       \\
    Reactive (scratch)        &     $ 68.0  \pm 1.3$ &     $  65.8 \pm 1.4$ &     $  65.7 \pm   1.5$  &     $ 5.1  \pm 0.3$ &     $ 5.3  \pm 0.2$ &     $ 5.3  \pm 0.2$ &     $6.2  \pm 0.0$ &     $6.0 \pm  0.0$ &     $6.0  \pm 0.0$ \\
    RNN (scratch)             &     $ 79.0  \pm 2.0$ &     $  74.0 \pm 0.8$ &     $  73.4 \pm   1.3$  &     $ 5.9  \pm 0.0$ &     $ 5.9  \pm 0.3$ &     $ 6.0  \pm 0.2$ &     $6.0  \pm 0.0$ &     $5.9 \pm  0.0$ &     $5.9  \pm 0.0$ \\
    SMT (scratch)             &     $104.8  \pm 2.2$ &     $ 101.6 \pm 0.9$ &     $  99.2 \pm   2.9$  &     $ 6.9  \pm 0.6$ &     $ 6.6  \pm 0.2$ &     $ 7.4  \pm 0.2$ &     $7.0  \pm 0.2$ &     $6.8 \pm  0.1$ &     $6.7  \pm 0.1$ \\
    SMT (MidLevel)            &     $155.7  \pm 2.0$ &     $ 145.1 \pm 2.3$ &     $ 134.2 \pm   1.8$  &     $10.6  \pm 0.3$ &     $10.6  \pm 0.6$ &     $10.8  \pm 0.4$ &     $7.6  \pm 0.2$ &     $7.3  \pm 0.1$ &     $7.3  \pm 0.2$ \\
    SMT (Video)               &     $141.2  \pm 4.5$ &     $ 129.2 \pm 2.1$ &     $ 125.8 \pm   2.6$  &     $10.8  \pm 0.6$ &     $10.0  \pm 0.4$ &     $ 9.6  \pm 0.1$ &     $7.5  \pm 0.1$ &     $7.4 \pm  0.0$ &     $7.4  \pm 0.0$ \\
    OccupancyMemory           &     $155.6  \pm 1.4$ &     $  86.6 \pm 2.2$ &     $  85.2 \pm   2.4$  & $\bm{14.1} \pm 0.6$ &     $10.9  \pm 0.2$ &     $10.2  \pm 0.3$ &     $7.8  \pm 0.1$ &     $5.8  \pm 0.0$ &     $5.8  \pm 0.0$ \\
    EPC                       & $\bm{172.5} \pm 0.6$ &  $\bm{161.6}\pm 3.1$ &  $\bm{159.3}\pm   2.0$  &     $12.7  \pm 0.2$ & $\bm{12.0} \pm 0.8$ & $\bm{12.0} \pm 0.1$ & $\bm{9.0} \pm 0.1$ & $\bm{8.5} \pm 0.1$ & $\bm{8.5} \pm 0.3$ \\

    \midrule
                              &                                                                                  \multicolumn{9}{c}{\textbf{Gibson-S}}                                                                                 \\
                              &\multicolumn{3}{c}{Area coverage  (m$^2$)}                             &                  \multicolumn{3}{c}{Flee     (m)}               &           \multicolumn{3}{c}{Object cov. (\#cat.)}           \\
   \cmidrule(lr){2-4}\cmidrule(lr){5-7}\cmidrule(lr){8-10}
    Method                    &           NF         &            N-D       &           N-D,P         &     NF              &           N-D       &        N-D,P        &          NF        &           N-D      &        N-D,P       \\
    Reactive (scratch)        &     $ 17.4  \pm 0.2$ &     $  17.8 \pm 0.4$ &     $  17.8 \pm   0.4$  &     $ 1.9  \pm 0.1$ &     $ 1.8  \pm 0.1$ &     $ 1.8  \pm 0.1$ &          -         &           -        &          -         \\
    RNN (scratch)             &     $ 20.6  \pm 0.4$ &     $  21.5 \pm 0.3$ &     $  21.6 \pm   0.2$  &     $ 2.3  \pm 0.2$ &     $ 2.2  \pm 0.2$ &     $ 2.2  \pm 0.2$ &          -         &           -        &          -         \\
    SMT (scratch)             &     $ 23.0  \pm 0.7$ &     $  23.5 \pm 0.4$ &     $  23.4 \pm   0.4$  &     $ 3.3  \pm 0.2$ &     $ 3.3  \pm 0.0$ &     $ 2.8  \pm 0.1$ &          -         &           -        &          -         \\
    SMT (MidLevel)            &     $ 29.1  \pm 0.1$ &     $  30.8 \pm 0.4$ &     $  30.8 \pm   0.6$  &     $ 4.2  \pm 0.0$ &     $ 4.1  \pm 0.0$ &     $ 3.4  \pm 0.0$ &          -         &           -        &          -         \\
    SMT (Video)               &     $ 28.8  \pm 0.4$ &     $  30.8 \pm 0.4$ &     $  30.7 \pm   0.4$  &     $ 4.0  \pm 0.0$ &     $ 3.9  \pm 0.0$ &     $ 3.4  \pm 0.1$ &          -         &           -        &          -         \\
    OccupancyMemory           &     $ 29.4  \pm 0.0$ &     $  30.8 \pm 0.3$ &     $  30.6 \pm   0.2$  &     $ 2.8  \pm 0.0$ &     $ 3.1  \pm 0.0$ &     $ 3.0  \pm 0.2$ &          -         &           -        &          -         \\
    EPC                       & $ \bm{31.5} \pm 0.1$ &   $\bm{34.0}\pm 0.2$ &   $\bm{34.0} \pm  0.2$  & $ \bm{4.5} \pm 0.0$ &  $\bm{4.6} \pm 0.2$ &  $\bm{4.4} \pm 0.1$ &          -         &           -        &          -         \\
    \midrule
                              &                                                           \multicolumn{9}{c}{\textbf{Gibson-L}}                                                                             \\
                              &\multicolumn{3}{c}{Area coverage  (m$^2$)}                             &                  \multicolumn{3}{c}{Flee     (m)}               &           \multicolumn{3}{c}{Object cov. (\#cat.)}           \\
   \cmidrule(lr){2-4}\cmidrule(lr){5-7}\cmidrule(lr){8-10}
    Method                    &           NF         &            N-D       &           N-D,P         &     NF              &           N-D       &        N-D,P        &          NF        &           N-D      &        N-D,P       \\
    Reactive (scratch)        &     $ 22.8  \pm 0.6$ &     $  22.4 \pm 0.2$ &     $  22.4 \pm   0.2$  &     $ 2.5  \pm 0.3$ &     $ 2.6  \pm 0.4$ &     $ 2.6  \pm 0.4$ &          -         &           -        &          -         \\
    RNN (scratch)             &     $ 28.6  \pm 0.3$ &     $  27.9 \pm 2.4$ &     $  28.2 \pm   2.5$  &     $ 2.8  \pm 0.4$ &     $ 2.7  \pm 0.4$ &     $ 2.8  \pm 0.4$ &          -         &           -        &          -         \\
    SMT (scratch)             &     $ 32.3  \pm 0.8$ &     $  33.4 \pm 1.2$ &     $  32.6 \pm   1.9$  &     $ 4.4  \pm 0.4$ &     $ 4.6  \pm 0.2$ &     $ 4.4  \pm 0.1$ &          -         &           -        &          -         \\
    SMT (MidLevel)            &     $ 47.2  \pm 1.6$ &     $  49.2 \pm 0.4$ &     $  46.8 \pm   2.8$  &     $ 6.0  \pm 0.4$ &     $ 5.4  \pm 0.4$ &     $ 5.1  \pm 0.6$ &          -         &           -        &          -         \\
    SMT (Video)               &     $ 47.6  \pm 2.4$ &     $  48.3 \pm 1.5$ &     $  48.0 \pm   1.0$  &     $ 6.5  \pm 0.4$ &     $ 6.0  \pm 0.3$ &     $ 4.8  \pm 0.4$ &          -         &           -        &          -         \\
    OccupancyMemory           & $ \bm{67.4} \pm 0.9$ &     $  56.8 \pm 0.8$ &     $  56.9 \pm   0.8$  &     $ 7.0  \pm 0.4$ &     $ 6.9  \pm 0.4$ &     $ 6.9  \pm 0.3$ &          -         &           -        &          -         \\
    EPC                       &     $ 62.2  \pm 1.0$ &   $\bm{66.7}\pm 1.5$ &  $ \bm{64.2}\pm   1.8$  &  $\bm{8.0} \pm 0.5$ &  $\bm{8.2} \pm 1.0$ &  $\bm{8.2} \pm 0.6$ &          -         &           -        &          -         \\
    \bottomrule
    \end{tabular}
}
\caption{
    Comparing robustness to sensor noise on downstream tasks in Gibson and Matterport3D. Note: NF denotes noise
    free sensing, N-D denotes noisy depth (and noise-free pose), and N-D,P denotes noisy depth and pose.
}
\label{tab:appendix_noise_robustness}
\end{table*}

\section{OccupancyMemory performance on Flee}\label{appsec:occmem}
OccupancyMemory relies on a global policy that samples a spatial goal location for navigation. A local navigation policy~\cite{wijmans2019decentralized}
then executes a series of low-level actions to reach that goal. Our qualitative analyses indicate that the global policy overfit to large MP3D environments.
It often samples far away exploration targets, relying on the local navigator to explore the spaces along the sampled direction. However, this strategy
fails in the small Gibson-S environments (typically a single room). Selecting far away targets results in the local navigator oscillating in place trying
to exit a single-room environment. This does not affect area coverage much because it suffices to stand in the middle of a small room and look at all sides.

\end{document}